\documentclass[letterpaper, 10 pt, conference]{ieeeconf}  

\IEEEoverridecommandlockouts                              
\usepackage[T1]{fontenc} 
\usepackage{graphicx,subfig}

\usepackage{hyperref}
\usepackage{booktabs}  
\usepackage{threeparttable} 
\usepackage{makecell}
\usepackage{graphicx}
\usepackage{multicol}
\usepackage{multirow}
\usepackage{feyn}
\usepackage{amsmath,amssymb,amsfonts}
\usepackage{ulem}
\usepackage{caption}
\usepackage{fancyhdr} 
\usepackage{array}
\usepackage{microtype}

\usepackage{hyperref}
\hypersetup{
	colorlinks=true,
	linkcolor=cyan,
	filecolor=blue,      
	urlcolor=black,
	citecolor=green,
}

\usepackage[sorting=none, style=ieee]{biblatex}

\title{\LARGE \bf
Erase, then Redraw: A Novel Data Augmentation Approach for Free Space Detection Using Diffusion Model
}

\author{Fulong Ma, Weiqing Qi, Guoyang Zhao, Ming Liu, and Jun Ma, \textit{Senior Member, IEEE}   
\thanks{Fulong Ma, Weiqing Qi, Guoyang Zhao, and Ming Liu are with The Hong Kong University of Science and Technology (Guangzhou), Guangzhou, China. (email: \{fmaaf,wqiad,gzhao492\}@connect.hkust-gz.edu.cn, eelium@hkust-gz.edu.cn.)}%
\thanks{Jun Ma is with The Hong Kong University of Science and Technology (Guangzhou), Guangzhou, China, and also with The Hong Kong University of Science and Technology, Hong Kong SAR, China. (email: jun.ma@ust.hk).}%
}

\begin{document}

\newcommand{\etal}{\textit{et al.}} 

\maketitle
\thispagestyle{empty}
\pagestyle{empty}

\begin{abstract}

Data augmentation is one of the most common tools in deep learning, underpinning many recent advances including tasks such as classification, detection, and semantic segmentation. The standard approach to data augmentation involves simple transformations like rotation and flipping to generate new images. However, these new images often lack diversity along the main semantic dimensions within the data. Traditional data augmentation methods cannot alter high-level semantic attributes such as the presence of vehicles, trees, and buildings in a scene to enhance data diversity. In recent years, the rapid development of generative models has injected new vitality into the field of data augmentation. In this paper, we address the lack of diversity in data augmentation for road detection task by using a pre-trained text-to-image diffusion model to parameterize image-to-image transformations. Our method involves editing images using these diffusion models to change their semantics. In essence, we achieve this goal by erasing instances of real objects from the original dataset and generating new instances with similar semantics in the erased regions using the diffusion model (as shown in Fig. \ref{diagram}), thereby expanding the original dataset. 
We evaluate our method on the KITTI road dataset \cite{fritsch2013new} and the Cityscapes dataset \cite{cordts2016cityscapes}, and our method achieves the best results compared to other data augmentation methods on both datasets, which demonstrates superiority and effectiveness of our proposed method.
Here is our project page: \href{https://sites.google.com/view/data-augmentation}{https://sites.google.com/view/data-augmentation}. 

\end{abstract}

\section{INTRODUCTION}

In recent years, artificial intelligence has been rapidly advancing, and autonomous driving has emerged as one of the largest engineering applications within the field. It is also considered one of the most challenging areas to develop. For autonomous vehicles, similar to lane detection \cite{ma2024monocular}, free space detection is a fundamental component of driving scene understanding. Free space detection methods typically classify each pixel in RGB or depth images as belonging to a drivable area or non-drivable area. These pixel-level classification results are then utilized by other modules in the autonomous driving system, such as trajectory prediction and path planning \cite{thoma2019mapping}, to ensure that the autonomous vehicle can navigate safely in complex environments
\cite{ma2023self}.

\begin{figure}[t]
    \setlength{\abovecaptionskip}{0pt}
    \setlength{\belowcaptionskip}{0pt}
    \centering
    \includegraphics[width=1.0\linewidth]{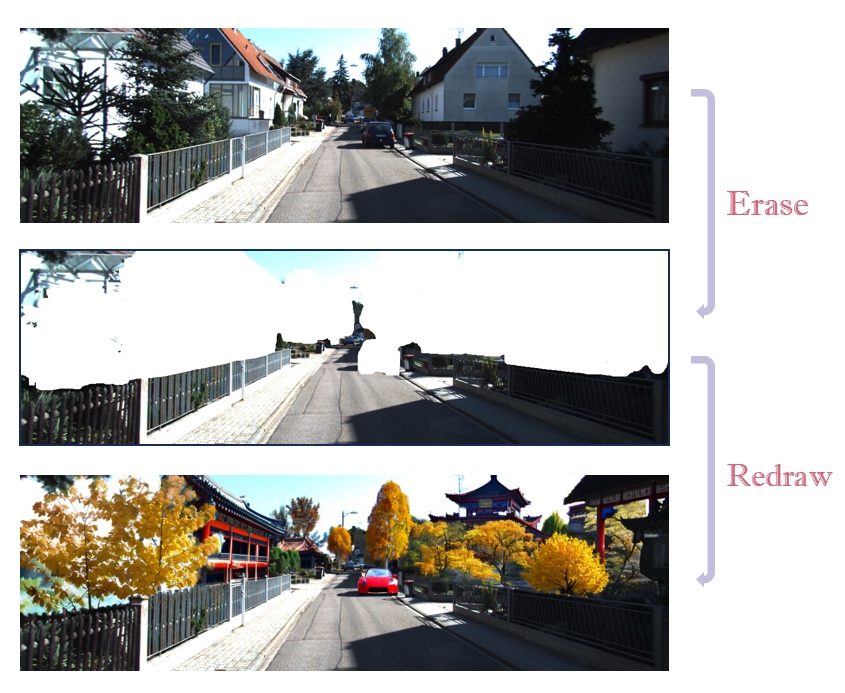}
    \captionsetup{font={footnotesize}}
    \caption{A schematic diagram of our data augmentation method, it involves first erasing the region of interest within the background of the original image, and then re-drawing within the erased area to generate new synthetic data.}
    \label{diagram}
\end{figure}

The current mainstream free space detection methods are mainly based on deep learning, which generally require a large amount of manually labeled data to train the algorithms. Manual data labeling is a costly, time-consuming, and labor-intensive task, which greatly affects the practical application of learning-based algorithms. In order to reduce the drawbacks of manual data labeling, researchers have proposed different solutions, including semi-supervised or self-supervised methods. For example, in \cite{han2018semisupervised}, a semi-supervised learning (SSL) method based on Generative Adversarial Networks (GANs) and a weakly supervised learning (WSL) method based on Conditional GANs (CGANs) was introduced. Compared to semi-supervised methods, self-supervised methods can further reduce the burden of data labeling. Mayr \textit{et al.} \cite{mayr2018self} proposed a self-supervised method that leverages the v-disparity image to automatically annotate training data for free space. Ma \textit{et al.} \cite{ma2023self} utilize depth information from LiDAR combined with road boundary detection to automatically generate training labels for free space on images.
In addition to semi-supervised and self-supervised approaches, data augmentation is also an attractive direction. By using various methods to generate more simulated data on a limited training dataset, the original dataset can be expanded to improve the performance of the model.

In this paper, we propose a novel data augmentation method for free space detection. The method is mainly divided into two steps: 
The method is mainly divided into two steps: first, using traditional instance segmentation algorithms (such as Mask R-CNN \cite{he2017mask}) or general segmentation algorithms (such as Segment Anything (SAM) \cite{kirillov2023segment}) to erase instance pixels in the background while keeping the foreground pixels unchanged.

Then, the erased regions are locally redrawn using a pre-trained diffusion model to restore the erased parts in the image, as shown in the Fig. \ref{diagram}. During the redrawing process, different linguistic prompts can be used to achieve redrawing of different objects and styles, providing great flexibility. We then test our proposed method on the KITTI road dataset \cite{fritsch2013new} and Cityscapes dataset \cite{cordts2016cityscapes}, and the experimental results demonstrate the effectiveness of our approach.
Our main contributions are as follows:
\begin{itemize}
\item We propose a novel data augmentation method specifically for the task of free space detection, which generates synthetic data through two steps of erasing background instances and redrawing. To the best of our knowledge, this is the first data augmentation method designed specifically for free space detection.
\item During the redrawing process, our method can adjust the objects' categories and styles of the redrawn areas through different text prompts. This distinguishes our method from previous data augmentation techniques and greatly enhances the flexibility of data augmentation.
\item We conduct comprehensive experiments on KITTI road dataset and Cityscapes dataset, and the results demonstrate that our data augmentation method achieves the best performance in the free space detection task.


\end{itemize}

\begin{figure*}[h]
    \setlength{\abovecaptionskip}{0pt}
    \setlength{\belowcaptionskip}{0pt}
    \centering
    \includegraphics[width=0.98\linewidth]{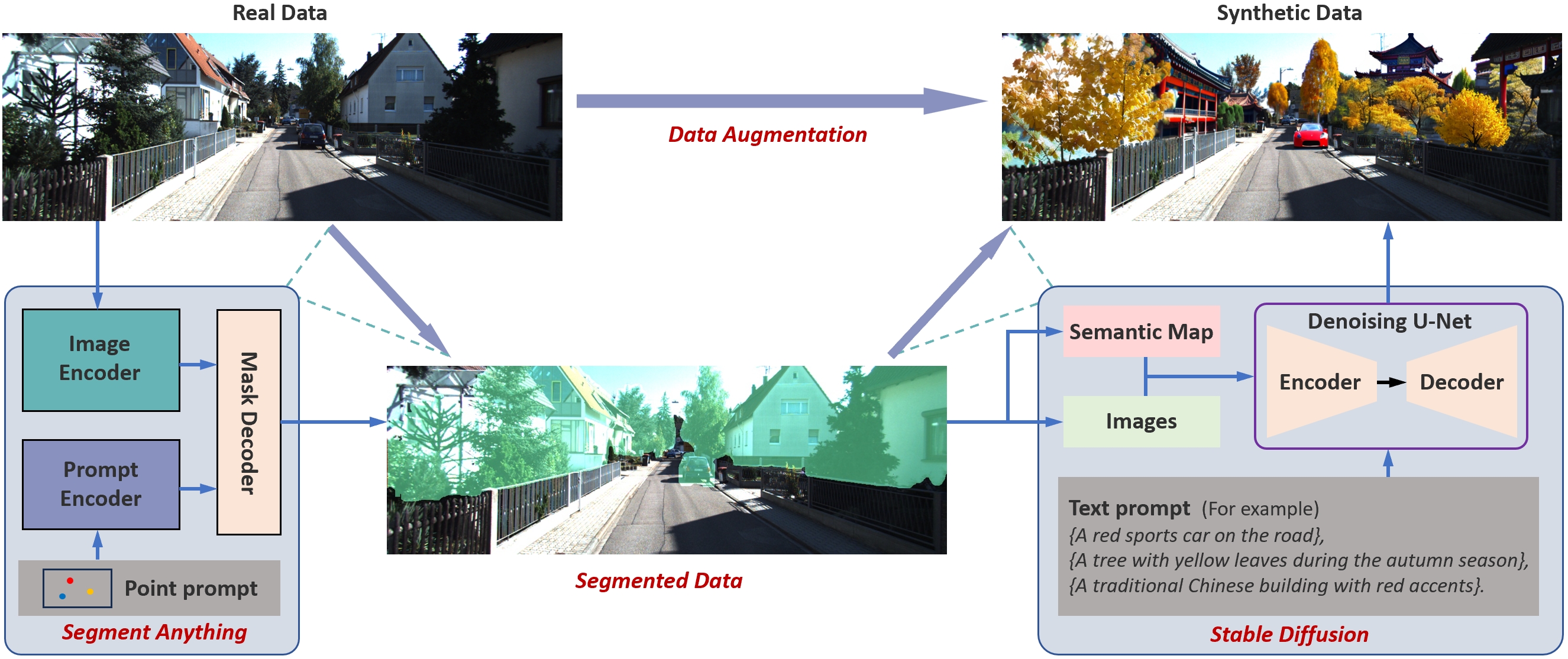}
    \captionsetup{font={footnotesize}}
    \caption{The architecture of our proposed data augmentation pipeline. Our pipeline consists of two parts, namely, SAM-based erasing and stable-diffusion-based scene redrawing.}
    \label{pipeline}
\end{figure*}

\section{RELATED WORKS}

\subsection{Free Space Detection}

Free space detection is generally divided into image-based methods, point cloud-based methods, and multimodal methods. In image-based methods, they can be further divided into methods based on the front view and methods based on Bird's Eye View (BEV). In image-based methods, there are methods that detect obstacles in column pixels \cite{levi2015stixelnet} to obtain free space, as well as methods based on semantic segmentation \cite{liu2018segmentation}. In point cloud-based methods, they can be divided into traditional methods and deep learning-based methods. In traditional methods, the free space is usually determined based on the spatial structure information of the point cloud through geometric rules \cite{narksri2018slope,zermas2017fast,himmelsbach2010fast,lee2022patchwork++}. Learning-based methods include projecting point clouds onto a spherical surface, converting them into sphere images for use with 2D convolution methods \cite{wu2019squeezesegv2}, as well as methods that directly take point clouds as input for deep neural networks \cite{qi2017pointnet}. To fully utilize the information from multiple sensors, researchers have developed multimodal fusion methods \cite{fan2020sne,chen2019progressive,chang2203fast}, to improve algorithm performance. PLARD \cite{chen2019progressive} first converts point clouds into ADI images, then inputs the ADI images together with RGB images into a deep neural network for end-to-end learning. SNE-RoadSeg \cite{fan2020sne} integrates normal information and image information to detect free space, while USNet \cite{chang2203fast} utilizes RGB images and binocular depth images combined with uncertainty estimation to achieve precise and efficient free space detection.

\subsection{Diffusion Model}
The diffusion model is a borrowed concept from thermodynamics, originating from the phenomenon of diffusion. In the field of statistics, this term refers to the process of transforming complex distributions into simpler distributions. In artificial intelligence, the diffusion model \cite{yang2023diffusion} defines a probabilistic distribution transformation model, where the forward propagation process can transform a complex distribution into a standard normal distribution.
Currently, the diffusion model has achieved significant applications in multiple fields. For image generation task, Stable Diffusion \cite{rombach2022high} can generate high-quality picture from noise under the guidance of text prompts. This has wide application prospects in areas such as art creation and game design.
Text generation \cite{li2023diffusion}, By training the diffusion model to learn the distribution of text data, we can generate text content with a certain semantic coherence. This has important application value in natural language processing, machine translation, and other fields.
Data augmentation \cite{trabucco2023effective}, In cases where the dataset is small or the annotation cost is high, we can use the diffusion model for data augmentation, generating more training samples to improve the model's performance.

\subsection{Data Augmentation}
Data augmentation aims to generate additional training data through certain methods to enhance model performance, including improving robustness, generalization ability, avoiding overfitting, and so on. Data augmentation can be divided into basic data augmentation and advanced data augmentation. 
In basic data augmentation methods, there are mainly three types: image manipulation, image erasing, and image mix. Image manipulations focus on image transformations, such as rotation, flipping, and cropping, etc \cite{buslaev2020albumentations}. Image erasing typically deletes one or more sub-regions in the image, with the main idea being to replace the pixel values of these sub-regions with constant values or random values \cite{chen2020gridmask}. Image Mix methods are mainly accomplished by mixing two or more images or sub-regions of images into one \cite{inoue2018data}.
In terms of advanced approaches, there are mainly three directions: auto augment, feature augmentation, and deep generative models. Auto augment is based on the fact that different data have different characteristics, so different data augmentation methods have different benefits \cite{cubuk2019autoaugment}. Rather than conducting augmentation only in the input space, feature augmentation performs the transformation in a learned feature space \cite{devries2017dataset}. The core idea of deep generative models is that the data distribution we generate data from should not be different from the original one, with GANs \cite{creswell2018generative} being one of the representative methods.

\section{METHOD}
\subsection{Preliminaries: Diffusion Model}
The diffusion probabilistic model was introduced in  \cite{ho2020denoising}, abbreviated as diffusion model. This is the pioneering work that applied the diffusion model to the field of image generation. Diffusion model is a Markov chain that includes both a forward process with a specific expression, and a backward process that is learned using neural networks. 
For the forward process, for an image $x_{0}$, apply a forward diffusion Markov process to add noise to the image over multiple time steps $t$ with a scheduled variance $\beta_{t}$:

\begin{equation}
    q(x_{t}|x_{t-1}) = \mathcal{N}(x_{t};\sqrt{1-\beta_{t}}x_{t}, \beta_{t}\textbf{I}),
\end{equation}
\begin{equation}
    q(x_{1:T} | x_{0}) = \prod_{t=1}^{T} q(x_{t}|x_{t-1}),
\end{equation}
where $T$ represents the complete set of steps. As $T$ approaches infinity, the resulting output will tend to an pure Gaussian distribution. 
Through the Markov process, we can calculate $x_{t}$ by:
\begin{equation}
    \begin{split}
     x_{t} & = \sqrt{\alpha_{t}} x_{t-1} + \sqrt{1 - \alpha_{t}} \epsilon_{t-1} \\
           & = \sqrt{\Bar{\alpha}_{t}} x_{0} + \sqrt{1 - \Bar{\alpha}_{t}} \epsilon ,
\end{split}
\end{equation}
where $\alpha_{t} = 1 - \beta_{t}$, $\Bar{\alpha}_{t} = \prod^{t}_{i=1}\alpha_{i}$, $\epsilon_{t-1}$, $\epsilon$ $\sim$ $ \mathcal{N}(0, \textbf{I})$.
The forward process is the process of adding noise, while the reverse process is the denoising process. If we can gradually obtain the reversed distribution $q(x_{t-1} | x_{t})$, we can reconstruct the original image distribution $x_{0}$ from the Gaussian distribution. It has been demonstrated that if $q(x_{t} | x_{t-1})$ satisfies a Gaussian distribution and $\beta$ is small enough, $q(x_{t-1} | x_{t})$ remains a Gaussian distribution. However, $q(x_{t-1} | x_{t})$ is unknown, so we use a deep neural network $p_{\theta}$ to approximate this distribution:
\begin{equation}
    p_{\theta}(x_{t-1} | x_{t}) = \mathcal{N}(x_{t-1}; \mu_{\theta}(x_{t},t),\Sigma_{\theta}(x_{t},t)),
\end{equation}
where $\mu_{\theta}(x_{t},t) = \frac{1}{\sqrt{\alpha_{t}}}(x_{t} - \frac{\beta_{t}}{\sqrt{1 - \Bar{\alpha}_{t}}}\epsilon_{\theta}(x_{t},t))$. The loss function for training the diffusion model:
\begin{equation}
    \mathcal{L} = \left [ \mathbb{E}_{t,x_{0},\epsilon_{t}}||\epsilon_{t} - \epsilon_{\theta}(\sqrt{\Bar{\alpha}_{t}}x_{0} + \sqrt{1 - \Bar{\alpha}_{t}}\epsilon_{t}, t)||^{2} \right ].
\end{equation}

At inference time, we start from a random noise $x_{T} \sim \mathcal{N}(0,\textbf{I})$, and then iteratively apply the model $\epsilon_\theta$ to obtain $x_{t-1}$ from $x_{t}$ until $t = 0$.

\subsection{Our Approach}
In this section, we will introduce our novel ``Erase, then Redraw” data augmentation method, and the overall process is shown in Fig. \ref{pipeline}.


Most previous work using generative models for data
augmentation focuses on classification, where each sample
is assigned a label from a finite set of possible classes. While semantic segmentation can be formulated as a classification
task in which each pixel is assigned a class, it introduces
an additional difficulty, namely that the position of the
objects matters.
Existing data augmentation methods primarily involve erasing parts of an image and filling them with black pixels or using parts of other images to fill in the erased regions. The result of these methods is that the generated data disrupts the original vision structure. 
Although this may enhance algorithm performance, the generated data are quite bizarre and would never be encountered in reality.

Fortunately, with the powerful image generation algorithm like diffusion models, we propose to utilize the shape information of objects in the image background and text prompts to generate higher quality synthetic data for data augmentation.
An image $x$ to be augmented contains masks $ \{ m_{i} \}_{i=1}^{N} $, where $N$ is the number of masks and each masked region $ x\oplus m_{i} $ contains only one object. For each image-mask pair, we also have a corresponding text prompt $p_{i}$, like ``a sports car on the road”.

\begin{figure}[h]
    \setlength{\abovecaptionskip}{0pt}
    \setlength{\belowcaptionskip}{0pt}
    \centering
    \includegraphics[width=1.0\linewidth]{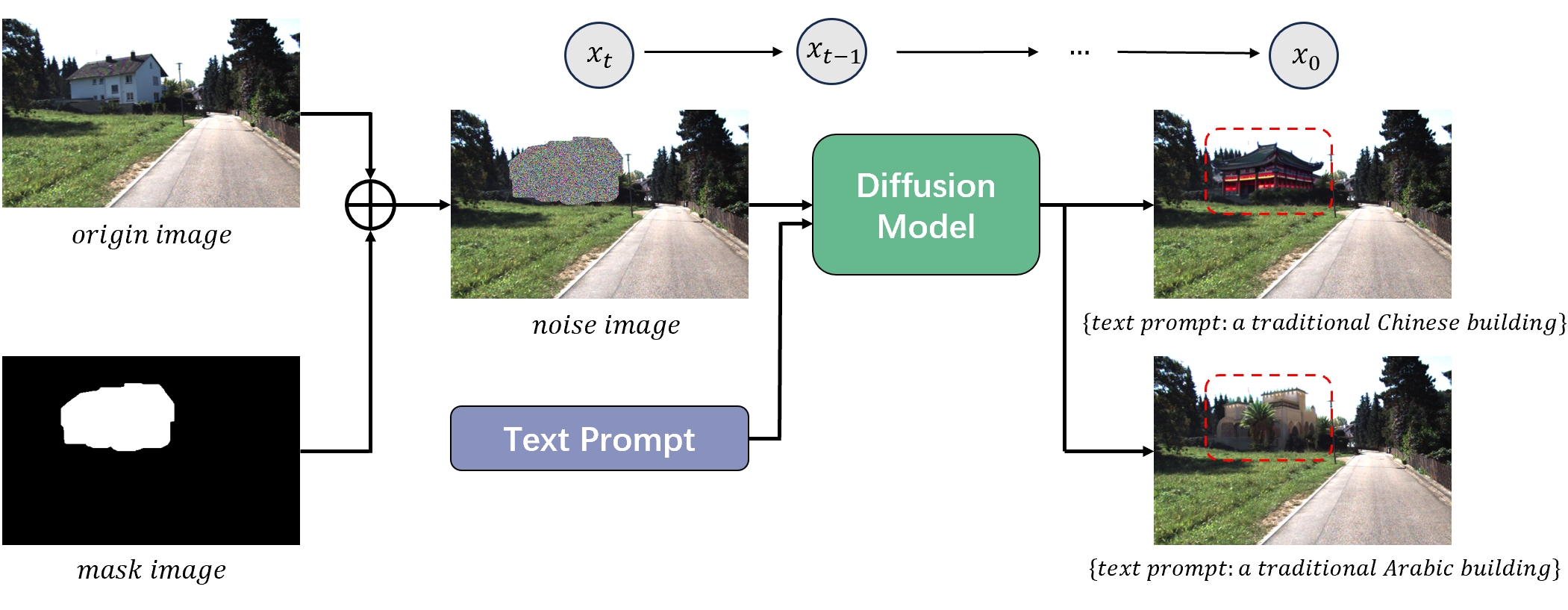}
    \captionsetup{font={footnotesize}}
    \caption{
    The redrawing process of our method. After erasing the pixels of the region of interest, new data is generated through the reverse diffusion process of the well-trained diffusion model. Different text prompts can generate new image with different distributions. For example, in the figure, our textual prompts are ``\textit{a traditional Chinese building}” and ``\textit{a traditional Arabic building}”, resulting in the erased area producing buildings with completely different architectural styles.}
    \label{denoise}
\end{figure}

\begin{figure*}[h]
    \setlength{\abovecaptionskip}{0pt}
    \setlength{\belowcaptionskip}{0pt}
    \centering
    \includegraphics[width=1.0\linewidth]{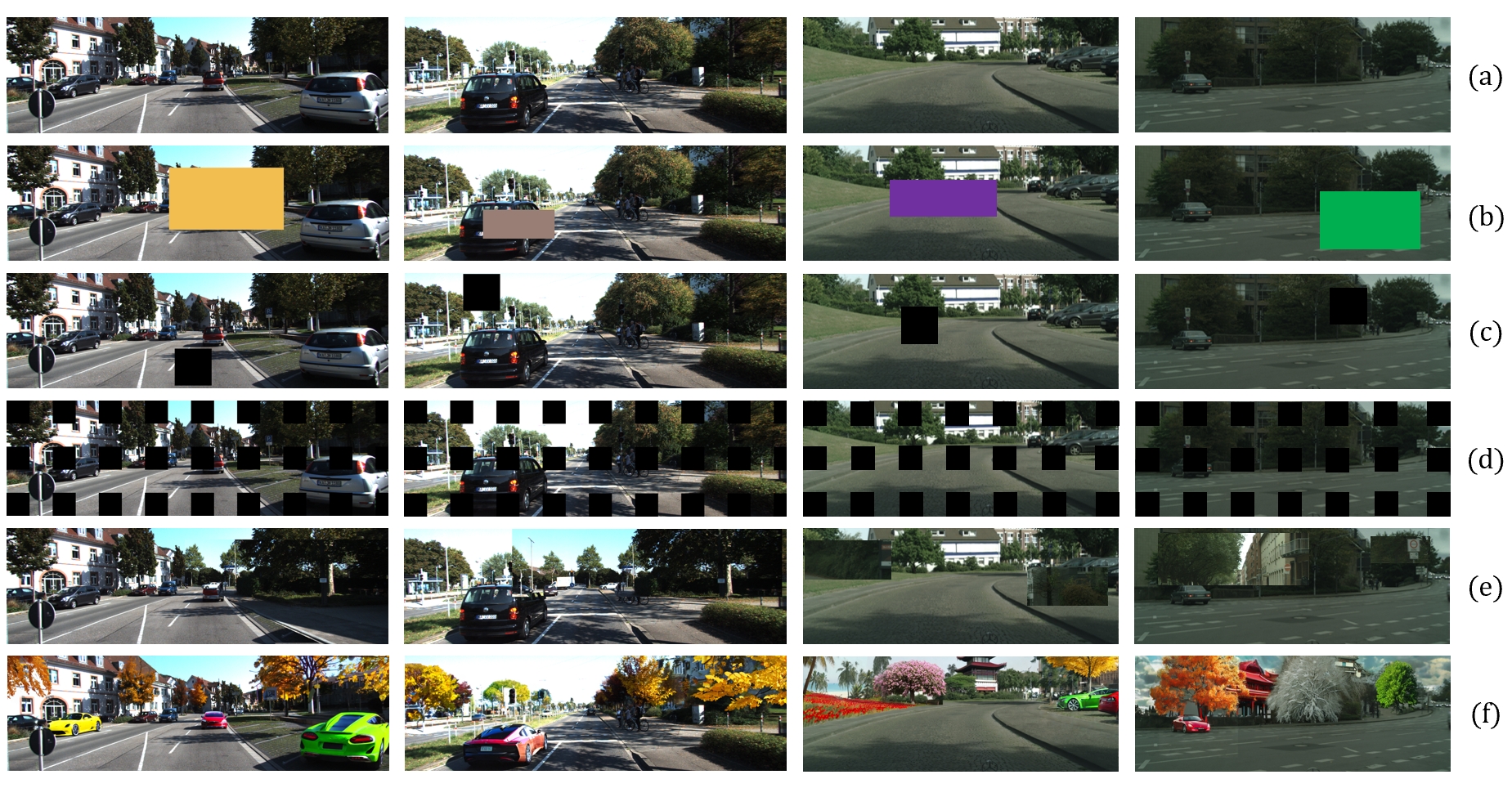}
    \captionsetup{font={small}}
    \caption{
    The comparison between the synthetic data generated by our method and the synthetic data generated by other data augmentation methods. The first row of images represents the original data from the KITTI road dataset \cite{fritsch2013new}. The 2nd, 3rd, 4th, 5th, and 6th rows correspond to the synthetic data generated by the data augmentation methods RandomErasing \cite{zhong2020random}, Cutout \cite{devries2017improved}, Gridmask \cite{chen2020gridmask}, CutMix \cite{yun2019cutmix}, and our method, respectively.
}
    \label{examples}
\end{figure*}

In the forward process, we select a segmentation mask $m_{i}$ from the background for image $x$ and its corresponding text prompt $p_{i}$. In our setup, $x_{0} = x$ and we add noise only to the pixels within the masked area, not to all pixels.
\begin{equation}
    \tilde{x_{t}} = \sqrt{\Bar{\alpha}_{t}} x_{0} + \sqrt{1 - \Bar{\alpha}_{t}} \epsilon,
\end{equation}
\begin{equation}
    x_{t}  = \tilde{x_{t}} \oplus m + x_{0} \oplus (1 - m),
\end{equation}
where $\epsilon \sim \mathcal{N} (0, \textbf{I})$ and $t$ represents the timestep in the forward process, $x_{t}$ represents the the image where the mask area is filled with Gaussian noise, as shown in the ``\textit{noise image}” of Fig. \ref{denoise}. We then use the $x_{t}$ and text prompt $p_{i}$ as input to the diffusion model so it can utilize the input and clear background information to restore the masked area $x_{0} \oplus m_{i}$. The whole process of this procedure is shown in Fig. \ref{denoise}.

For the ``erasing” process, \textit{i.e.}, the masks $ \{ m_{i} \}_{i=1}^{N} $ generation step, we utilized SAM \cite{kirillov2023segment} model, proposed by Menta AI. 
As a foundational model for image segmentation, SAM demonstrates strong generalization capabilities and performs well across different data domains. 
We utilize the method proposed in \cite{li2023clip}, which involves using text prompts specifying the objects to be erased to generate candidate points on the original image. These candidate points are then used as point prompts for the SAM model to erase objects in the background of the original image. Please note that when erasing background instance pixels, we ensure that foreground pixels remain unaffected, thereby avoiding inconsistencies between the training labels of the generated synthetic data and the original data.
As shown in row (f) of Fig. \ref{examples}, the foreground pixels (free space) of each synthetic data sample remain consistent with the original image. We only erase and redraw elements in the background to alter the original data distribution, thereby achieving the effect of data augmentation.


For the ``redrawing” process, \textit{i.e.}, the reverse diffusion process, we used the well-trained Stable Diffusion \cite{rombach2022high} model. Stable Diffusion can generate high-quality realistic simulation data based on text prompts. The input for Stable Diffusion can be text prompts for generating images from text, or it can be an image along with text prompts, used to adapt the input image based on the text prompt. In our proposed method, our input consists of image and text prompts. By using diverse text prompts (a text prompt generator\footnote{https://socialbu.com/tools/generate-prompt-text2img} can be used to conveniently generate high-quality text prompts), we are able to generate a wide variety of synthetic data in different styles, making it very flexible and capable of generating more diverse augmentation data. The synthetic data generated by our method is shown in the last row of Fig. \ref{examples}.

\begin{table}[h]
\centering
\captionsetup{font={footnotesize}}
\renewcommand\arraystretch{1.0}
\caption{Description of baseline methods.}
\label{baseline}
\begin{tabular}{m{1.8cm}<{\centering}|m{5.6cm}<{\centering}}
\toprule
Methods & \makecell[c]{ Description }   \\
\hline

Standard &  \scriptsize No data augmentation is applied, with the algorithm being trained solely on the original dataset provided.\\
\hline
Basic & \scriptsize Dataset undergoes a sequence of transformations, including horizontal flipping, random rotation, brightness and contrast adjustments, and elastic deformation. \\
\hline
RandomErasing & \scriptsize Randomly
selects a rectangle region in an image and erases its pixels with random values.\\
\hline
Cutout&\scriptsize Cutout employs a fixed-size square area, filled entirely with 0 (black), and permits the square area to extend outside the image. \\
\hline
CutMix&\scriptsize Randomly select two images, and randomly crop a rectangular area from each image. Then, exchange the cropped areas between the two images and merge them into a new image.
\\
\hline
GridMask&\scriptsize Generate a structured grid array first, and then erase the image information within the grid cells.
 \\
\toprule
\end{tabular}
\end{table}

\begin{table*}[h]
\renewcommand\arraystretch{1.0}
\captionsetup{font={footnotesize}}
\caption{
The experimental results of our data augmentation method on the KITTI road dataset, as well as other data augmentation methods such as Basic \cite{buslaev2020albumentations}, RandomErasing \cite{zhong2020random}, Cutout \cite{devries2017improved},  CutMix \cite{yun2019cutmix}, and GridMask \cite{chen2020gridmask}. To ensure comprehensive experimentation, experiments were conducted on three different classic model on three different network architectures. Bold indicates the best result, while underline indicates the second-best result.
}

\label{table_1}
\begin{center}
\setlength{\tabcolsep}{4mm}{
\begin{tabular}{c c c c c c c c} 
\toprule
Network  & Network Architecture    & Augmentation Method     & Accuracy &Precision &Recall & F1-Score& mIoU \\

\hline
\multirow{6}{*}{U-Net \cite{ronneberger2015u}} &\multirow{6}{*}{CNN} & Standard &94.78 &82.74 & 87.50 &85.05 & 75.42 \\
 & &Basic       &95.19 &85.77  &87.63  &86.70  &76.52  \\
 & & RandomErasing &94.61 &79.43  &\underline{94.23}  &86.20  & 75.75 \\
 & & Cutout    &\underline{95.76}  &\underline{88.76} &87.34  & \underline{88.05} & \underline{78.65}  \\
 & & CutMix    &95.50  &85.94  &89.45  &87.66  &78.04  \\
 & & GridMask  &91.89  &69.73  & \textbf{96.50}  &80.96  &68.02  \\
 &  & \textbf{Ours}     &\textbf{96.59} &\textbf{92.42} &88.14&\textbf{90.23} &\textbf{82.20} \\
\hline
\multirow{6}{*}{Swin-UNet \cite{cao2022swin}} &\multirow{6}{*}{Transformer} & Standard &93.52 &80.60 &83.93 &82.23 &69.83 \\
 & &Basic      &94.76 &84.35  &86.98  &85.59  &74.82 \\
 & & RandomErasing &95.18 &84.41  &87.98  &86.15  &77.08  \\
 & & Cutout    &95.25  &83.36  &88.90  &86.04  &\underline{77.65}  \\
 & & CutMix    &95.20  &84.51  &\underline{89.59}  &86.97  & 76.96 \\
 & & GridMask  &\underline{95.27}  &\textbf{85.72}  &88.27  &\underline{86.98}  & 76.96  \\
 &  & \textbf{Ours}& \textbf{95.54} & \underline{85.66} &\textbf{90.16} &\textbf{87.85} & \textbf{78.34} \\
\hline
\multirow{6}{*}{VM-UNet \cite{ruan2024vm}} &\multirow{6}{*}{Mamba} 
& Standard & 97.86 & 93.51 & 94.56 & 94.03 & 88.73  \\
 & &Basic   & 97.86 & 92.82 & 95.39 & 94.09 & 88.84 \\
 & & RandomErasing & 98.44 & 96.14 & 95.07 & 95.60 & 91.58 \\
 & & Cutout    & \underline{98.59} & \textbf{96.99} & 95.08 & \underline{96.03} & \underline{92.35} \\
 & & CutMix    & 98.43 & 95.60 & \underline{95.61} & 95.60 & 91.57 \\
 & & GridMask  & 97.61 & 93.12 & 93.51 & 93.32 & 87.47 \\
 & & \textbf{Ours}& \textbf{98.65} & \underline{96.21} &\textbf{96.23} &\textbf{96.22} & \textbf{92.72} \\

\toprule
\label{kitti_experiment}
\end{tabular}}
\end{center}
\end{table*}

\begin{table*}[h]
\renewcommand\arraystretch{1.0}
\captionsetup{font={footnotesize}}
\caption{
The experimental results of our data augmentation method on the Cityscapes dataset, as well as other data augmentation methods such as Basic \cite{buslaev2020albumentations}, RandomErasing \cite{zhong2020random}, Cutout \cite{devries2017improved}, CutMix \cite{yun2019cutmix}, and GridMask \cite{chen2020gridmask}.  Bold indicates the best results, and underline indicates the second-best results.
}

\label{table_2}
\begin{center}
\setlength{\tabcolsep}{4mm}{
\begin{tabular}{c c c c c c c c} 
\toprule
Network  & Network Architecture    & Augmentation Method     & Accuracy &Precision &Recall & F1-Score& mIoU \\

\hline
\multirow{6}{*}{U-Net \cite{ronneberger2015u}} &\multirow{6}{*}{CNN} & Standard &93.89 &89.07 &94.23  &91.57 &84.21  \\
 & &Basic       &95.24 &89.35   &96.90   & 92.97 & 86.87 \\
 & & RandomErasing &96.35 &90.22 &\textbf{97.35}  &93.65  &87.33\\
 & & Cutout    &\underline{96.59}  &\underline{91.21} &96.88 & \underline{93.96} & \underline{88.54}  \\
 & & CutMix    &96.42 &90.87 &96.78  & 93.73 & 88.43 \\
 & & GridMask  &96.31  & 90.24 &95.99 & 93.03 & 87.98 \\
 &  & \textbf{Ours} &\textbf{97.01} &\textbf{92.38} &\underline{97.10} & \textbf{94.68}&\textbf{89.11} \\
\hline
\multirow{6}{*}{Swin-UNet \cite{cao2022swin}} &\multirow{6}{*}{Transformer} & Standard &94.31 &87.25 &95.17 &91.03 & 84.22\\
 & &Basic      &95.40 &89.56 &97.18  &93.21  &87.29 \\
 & & RandomErasing &95.88 &90.35  & \textbf{97.98} &94.01  & 88.34\\
 & & Cutout    &96.42  &91.26  &97.88  & 94.45 &88.97  \\
 & & CutMix    &\underline{96.47}  &91.15  &97.21  &94.08 &88.35  \\
 & & GridMask  &96.13  &90.98  &97.35  & 94.06 & 87.98  \\
 &  & \textbf{Ours}&\textbf{96.89} & \textbf{91.98} &\underline{97.91} & \textbf{94.85}& \textbf{89.35} \\
\hline
\multirow{6}{*}{VM-UNet \cite{ruan2024vm}} &\multirow{6}{*}{Mamba} 
& Standard & 95.16 &93.79  & 95.08 & 94.43 &89.77   \\
 & &Basic   &96.95 &94.28  &96.46  &95.36  &91.13  \\
 & & RandomErasing &97.87  &\underline{96.12}  &97.24  & \underline{96.67} &92.53 \\
 & & Cutout    &97.85  &96.03 &96.98 & 96.50 &\underline{92.99}  \\
 & & CutMix    &\underline{98.15} &95.78  & \underline{97.33} &96.55 & 92.87\\
 & & GridMask  &97.98  &96.00  &97.23  & 96.61 &92.56 \\
 & & \textbf{Ours}& \textbf{98.55} &\textbf{96.77} &\textbf{97.51} &\textbf{97.14} & \textbf{93.15} \\

\toprule
\label{table_single_modal}
\end{tabular}}
\end{center}
\end{table*}

\section{EXPERIMENT}

\subsection{Baselines}

In our experiments, we use the following data augmentation methods for comparative experiments: Standard, Basic, RandomErasing \cite{zhong2020random}, Cutout \cite{devries2017improved}, Cutmix \cite{yun2019cutmix}, and Gridmask \cite{yun2019cutmix}. Standard represents no data augmentation, where the algorithm is solely trained with the provided dataset, which is randomly split into training and validation sets. Basic represents data augmentation by using the Albumentations library \cite{buslaev2020albumentations}, which is also the simplest, most basic, and most commonly used data augmentation method.
The visualization of these data augmentation methods are shown in Fig. \ref{examples}.
Similar to our method, DA-Fusion \cite{trabucco2023effective} also employs diffusion model to generate simulated data. 
However, DA-Fusion is specifically designed for classification tasks, it is unsuitable for segmentation tasks.
Therefore it is not included in the comparison scope.
A summary of all baselines is presented in Table \ref{baseline}.
The augmented data generated using these methods and our approach is depicted in Fig. \ref{examples}.

\subsection{Dataset}
\textcolor{black}{
In our experiments, we use the KITTI road dataset \cite{fritsch2013new} and the Cityscapes dataset \cite{cordts2016cityscapes} to validate the effectiveness of our algorithm. The KITTI road dataset is one of the most popular and widely used datasets for road scene understanding, commonly employed for tasks such as free space detection and lane line detection. This dataset comprises 289 frames of training data and 290 frames of testing data. The Cityscapes dataset, on the other hand, focuses on semantic segmentation, instance segmentation, and panoptic segmentation tasks in urban street scenes, encompassing a total of 30 categories. It includes 2975 training images, 500 validation images, and 1525 testing images. Since our work is concentrated on the task of free space detection, we retained only the "road" category from the Cityscapes dataset to validate our method.
Since the test labels for both datasets are not public, we partition the original datasets. For the KITTI road dataset, we split the training data into 144 images for testing and 145 images for further division: 20\% as validation and the remaining 116 for training. For Cityscapes, we randomly select 50\% of the 2975 training images for training and the rest for testing.
}

\subsection{Experiment Setup}

Our experiments are conducted in an Ubuntu 20.04 environment, equipped with an Intel i7 12700F CPU and a NVIDIA GeForce RTX 4090 GPU. We employed the PyTorch framework for model training and set training parameters with a batch size of 2, a total of 300 epochs.
Regarding to the augmented data, we used each data augmentation method to generate 3 synthetic images for each origin image in datasets for training. In the standard experimental setup without any data augmentation, we duplicated each original image three times to maintain fairness in the amount of training data.

\subsection{Evaluation Metrics}

Consistent with other free space detection works, we selected five commonly used evaluation metrics to assess the performance of our proposed method. These evaluation metrics are:
$Accuracy$, $Precision$, $Recall$, $F_{Score}$ and $IoU$ (intersection over union),
and they were computed as follows: $Accuracy =   \frac{N_{TP} + N_{TN}}{N_{TP} + N_{FP}  + N_{TN}  + N_{FN}}, Precision =  \frac{N_{TP}}{N_{TP}+N_{FP}}, Recall =  \frac{N_{TP}}{N_{TP} + N_{FN}}, F1-Score =  \frac{2 * Precision * Recall}{Precision + Recall}, IoU =  \frac{N_{TP}}{ N_{TP} + N_{FP} + N_{FN}},$ where $N_{TP}$, $N_{TN} $, $N_{FP}$ and $N_{FN}$ represents the true positive, true negative, false positive, and false negative pixel numbers, respectively. 

\subsection{Performance Evaluation}


The quantitative experimental results of three different architectures of single-modal algorithms U-Net, Swin-UNet, and VM-UNet on the KITTI road dataset are shown in Table \ref{kitti_experiment}. From the table, it can be seen that our data augmentation method achieves the best performance on three different deep neural network architectures: CNN, Transformer, and Mamba. 
Specifically, compared to the second best method, 
on U-Net, our data augmentation method increased the F1-Score from 88.05 to 90.23 and mIoU from 78.65 to 82.20. 
On Swin-Net, our data augmentation method increased the F1-Score from 86.98 to 87.85 and mIoU from 77.65 to 78.34. 
On VM-UNet, our data augmentation method improved the F1-Score from 96.03 to 96.22 and mIoU from 92.35 to 92.72.
When compared to the Basic method, 
the F1-Score and mIoU increased by 2.26\% and 8.87\% on Unet, the F1-Score and mIoU increased by 2.64\% and 4.71\% on Swin-Unet, and the F1-Score and mIoU increased by 2.05\% and 3.94\% on VM-Unet.
\textcolor{black}{The quantitative experimental results on Cityscapes dataset \cite{cordts2016cityscapes} are shown in Table \ref{table_2}.
Similar to the results on the KITTI road dataset, we also achieve the best performance on the Cityscapes dataset.}


Based on the experiments conducted on KITTI road dataset and Cityscapes dataset, our data augmentation method has shown promising improvements, demonstrating the effectiveness of our approach.

\section{CONCLUSIONS}
\label{conclusions}

In this paper, we propose a novel data augmentation method for free space detection task using SAM model and diffusion models. 
Our method consists of two steps. First, we utilize a SAM to erase elements from the original data and retain pixel regions belonging to free space. Second, we deploy a pretrained diffusion model to inpaint the erased regions, allowing us to generate diverse and personalized synthetic data by leveraging language prompts. 
We tested our method on the KITTI road dataset, and the results demonstrate that our data augmentation approach achieves leading performance compared to existing methods. However, our method has a few limitations, and there are several directions for future work. 
Firstly, our method does not explicitly control how the diffusion model enhances images. Introducing a control mechanism, like the idea of ControlNet \cite{zhang2023adding}, in future work could better manage the generation of images in erased regions, potentially improving results. 
Secondly, expanding the data augmentation method presented in this paper to more vision tasks to enhance its versatility is also a direction worth exploring.

\normalem
\renewcommand*{\bibfont}{\footnotesize}
\printbibliography
\end{document}